\pdfoutput=1

\documentclass[11pt]{article}

\usepackage{emnlp2021}

\usepackage{times}
\usepackage{latexsym}

\usepackage[T1]{fontenc}

\usepackage[utf8]{inputenc}

\usepackage{graphicx}
\usepackage{caption}
\usepackage{amsmath}
\usepackage{amssymb}
\usepackage{pgfplots}
\pgfplotsset{compat=newest}
\usepackage{comment}
\usepackage{booktabs}
\usepackage{hyperref}
\usepackage[overload]{empheq}
\usepackage{makecell}
\usepackage{enumitem}
\usepackage{url}

\newcommand{\ella}[1]{{\textit{\color{blue}{#1}}}}
\newcommand{\ofer}[1]{{\textit{\color{purple}{#1}}}}
\newcommand{\segev}[1]{{\textit{\color{green}{#1}}}}

\title{We've had this conversation before: A Novel Approach to \\ Measuring Dialog Similarity}

\author{
    Ofer Lavi\hspace{1.7cm}
	Ella Rabinovich\hspace{1.7cm}
	Segev Shlomov\hspace{1.7cm}
	David Boaz
	\vspace{0.1cm} \\
	\textbf{Inbal Ronen}\hspace{4cm} 
	\textbf{Ateret Anaby-Tavor} 
	\vspace{0.15cm} \\
	IBM Research \\
	\texttt{\{oferl, davidbo, inbal, atereta\}@il.ibm.com} \\
	\texttt{\{ella.rabinovich1, segev.shlomov1\}@ibm.com}
}

\begin{document}
\maketitle
\begin{abstract}
Dialog is a core building block of human natural language interactions. It contains multi-party utterances used to convey information from one party to another in a dynamic and evolving manner. The ability to compare dialogs is beneficial in many real world use cases, such as conversation analytics for contact center calls and virtual agent design.

We propose a novel adaptation of the edit distance metric to the scenario of dialog similarity. Our approach takes into account various conversation aspects such as utterance semantics, conversation flow, and the participants. We evaluate this new approach and compare it to existing document similarity measures on two publicly available datasets. The results demonstrate that our method outperforms the other approaches in capturing dialog flow, and is better aligned with the human perception of conversation similarity. 

\end{abstract}

\section{Introduction}
\label{sec-introduction}

Measuring semantic textual similarity lies at the heart of many natural language and text processing tasks, such as sentence classification, information retrieval, and question answering. 
Traditional text representation approaches, such as  high dimensional and sparse feature vectors, have been boosted by the introduction of efficiently learned embeddings \citep{mikolov_efficient_2013, pennington-etal-2014-glove, bojanowski2017enriching}, unleashing the full power of the dense semantic representation of words. Subsequently, new methods were developed for contextual representation of words, sentences, paragraphs, and documents, facilitating the assessment of semantic similarity between larger portions of text \citep{le_distributed_2014, peters-etal-2018-deep, devlin-etal-2019-bert}. 

Conversations\footnote{Although not strictly equivalent, we use the terms \textit{conversation} and \textit{dialog} interchangeably hereafter.} differ from documents that compound multiple sentences or paragraphs in a number of key ways. They are semi-structured documents constructed from a sequence of utterances and they present unique characteristics such as having an author for each utterance, 
and a conversation flow that can be viewed as a skeleton built of \textit{dialog acts} \citep{mctear2016conversational}. 
Indeed, it has been shown that models adapted specifically to analyze conversations outperform those built to analyze general documents, when applied to dialog data \citep{wu_tod-bert_2020, ohsugi-etal-2019-simple, henderson-etal-2020-convert, zhang2020dialogpt}. A plausible assumption, therefore, would be that dialog similarity assessment could benefit from an approach adjusted specifically for this domain. 

Related work in this field is relatively sparse. 
\citet{appel_combining_2018} derive two different similarity functions between conversations, one for content similarity based on TF-IDF, and the other for the conversation structure, using engineered features related to the dialog flow. The scores, computed independently for the two dimensions, are further combined to infer the overall conversation similarity ranking. \citet{xu_clustering-based_2019} learn a distance function between utterances and conversations based on expert judgments and later use this function to cluster conversations. Their approach is specifically tailored for conversations with an automatic dialog system (i.e., bot), limiting its applicability to the much wider conversational domain.

In this work we draw on the concept of \textit{edit distance}, the family of metrics and algorithms \citep{wagner1974string} widely used for sequence analysis. This analysis is done mainly at the character level for strings or at the nucleotide or amino--acid level in computational biology \citep{navarro_guided_2001}. The edit distance has also been applied to sequences of sentences to detect the differences between documents \citep{zhang_sentence-level_2014, barzilay-elhadad-2003-sentence}.

We propose combining the power of edit distance in assessing sequence similarity with the power of distributional semantics to form a novel similarity measure between conversations. This new measure -- \texttt{convED} -- takes into account the utterance semantics, as well as the dialog flow and its unique traits. 
We suggest and evaluate a framework for seamless, non-intrusive, and elegant adaptation of the widely-used edit distance metric to the scenario of conversation similarity.\footnote{We release data annotated by crowd-workers, and used for evaluation at \url{https://ibm.biz/Bdfp3V}.}
The suggested approach can be practically leveraged for downstream applications, such as those in the domain of dialog pattern mining. 

\section{Model}
\label{sec-model}

In this section, we present a brief reminder of what the edit distance metric involves (Section~\ref{sec:min-edit-distance}), followed by the unique adaptations designed for the scenario of conversation similarity (Section~\ref{sec:conved}).

\subsection{Minimum Edit Distance}
\label{sec:min-edit-distance}

Our approach is inspired by the widely-used notion of sequence similarity -- \textit{edit distance}: the minimal number of insertions, deletions, and substitutions required to transform one sequence into another. Sequences are typically drawn from the same finite set of distinct symbols, e.g., the alphabet letters for strings. Given sequences $a$ and $b$ of lengths $m$ and $n$, the distance $d_{ij}$ between two arbitrary sequence prefixes -- of length $i$ and $j$, respectively -- is defined recursively by
\begin{subequations}
\label{eq:edit-distance}
    \begin{equation}
    \label{eq:edit-distance-1}
        d_{i,0} = \sum_{k=1}^{i} w_{del}(a_{k}), \; d_{0,j} = \sum_{k=1}^{j} w_{ins}(b_{k})
    \end{equation}
    \hspace{-1in}
    \begin{equation}
    \label{eq:edit-distance-2}
        d_{i,j} = min 
        \begin{cases} 
            d_{i-1, j} \hspace{10pt} + \hspace{15pt} w_{del}(a_{i}) \\
            d_{i, j-1} \hspace{10pt} + \hspace{15pt} w_{ins}(b_{j}) \\
            d_{i-1, j-1} + w_{sub}(a_{i}, b_{j}) \\
        \end{cases}
    \end{equation}
\end{subequations}

\noindent
for $i{\in}[1,m]$, $j{\in}[1,n]$, where $w_{del}$, $w_{ins}$ and $w_{sub}$ are deletion, insertion, and substitution weights, respectively; these vary  according to the precise application. The final edit distance between the two sequences $a$ and $b$ --- $d_{m,n}$ --- may then be computed using dynamic programming \citep{wagner1974string}. The chain of steps needed to convert one sequence into another constitutes the \textit{sequence alignment}, where each element in the first sequence is paired with an element or gap in the second one. As an example, one possible alignment between the words `shine' and `train' will result in following steps; assuming insertion and deletion cost of 1, and substitution cost of 2, the edit distance between these strings is 6.

\begin{table}[hbt]
\centering
\texttt{
\begin{tabular}{cccccc}
$\bullet$ & \textbf{s} & \textbf{h} & i & n & e \\
| & | & | & | & | & | \\
t & \textbf{r} & \textbf{a} & i & n & $\bullet$ \\
\end{tabular}
}
\end{table}

A dialog can essentially be viewed as a temporal organization of utterances, and their underlying dialog acts. In this context, a dialog act refers to a certain function, such as a request or statement. The unique nature of dialogs, as opposed to strings, poses unique challenges to the alignment procedure. We next describe these challenges, as well as the solutions we applied to address them. 

\subsection{Conversation Edit Distance (\texttt{convED})}
\label{sec:conved}
This work focuses on multi-party 
conversations in the domain of customer service, 
where a dialog is represented by an interaction between the actors: 
a customer and a customer support agent. Formally, for two conversations -- $c_1$ and $c_2$ of length $m$ and $n$ -- we define the sequence of utterances to be $u_1^1, ..., u_1^m$ and $u_2^1, ..., u_2^n$ produced by actors $a_1^1, ..., a_1^m$ for the first and $a_2^1, ..., a_2^n$ for the second conversation, respectively. 

\paragraph{Utterance Substitution Cost}
Motivated by the intuition that the alignment of two utterances -- $u_1^i$ and $u_2^j$ -- should be driven by their semantic similarity, we define the substitution cost of the two as a function of their distance in a semantic space. Namely, we encode $u_1^i$ and $u_2^j$ into distributional representations $e_1^i$ and $e_2^j$ using the Universal Sentence Encoder \citep{cer2018universal}.\footnote{Similar results were obtained when using the largest sentence-transformers model \citep{reimers-2019-sentence-bert}.} We define their substitution cost ($w_{sub}(u_1^i, u_2^j)$) as the \textit{cosine distance} of the representations, scaled by a factor $\alpha$ (see details on $\alpha$ optimization in Appendix A.1):
\begin{equation}
\label{eq:sub-cosine}
    w_{sub}(u_1^i, u_2^j) = \alpha{\times}(1\text{--}\cos(e_1^i, e_2^j))
\end{equation}

\paragraph{Alignment by Actor Type}
Structural similarity between two dialogs inherently implies a similarity between utterances matched by their actor type, whether customer or agent. Conversations in the customer support domain are likely to comprise a sequence of requests, clarification questions, solutions, actions, and confirmations. Dialogs that agree on the assignment of such patterns to actors would naturally be considered more similar than those that do not.
We impose \textit{inter-actor agreement} during the alignment process, by weighting the substitution of utterances produced by different actors with an infinitely high cost. Consequently, the algorithm will avoid making such cross-actor alignments. We re-define Equation~\ref{eq:sub-cosine}: 
\begin{equation*}
\centering
\fontsize{10}{12}{ 
\label{eq:sub-cosine-actors}
    w_{sub}(u_1^i, u_2^j) = \begin{cases} 
        \alpha{\times}(1\text{--}\cos(e_1^i, e_2^j)) \;\; \text{if } a_1^i == a_2^j \\
        \infty \hspace{15pt} \text{otherwise}
    \end{cases}
}
\end{equation*}



A sample invocation of the framework (with $w_{ins}$ and $w_{del}$ weights set to $1$) is presented in Table~\ref{tbl:conv-examples}, resulting in the utterance-level alignment of two dialogs in the domain of booking tickets.

\begin{table*}[hbt]
\centering
\resizebox{\textwidth}{!}{
\begin{tabular}{l|l|l}
\# & \multicolumn{1}{c}{\textbf{Conversation 1}}  & \multicolumn{1}{c}{\textbf{Conversation 2}} \\ \hline
1 & \makecell{\textbf{Customer}: I'd like to look for a film to watch. \\ I like adventure films.} & \textbf{Customer}: I'd like to search for a fun film to watch. \\ \hline
2 & \textbf{Agent}: Where are you located? & \textbf{Agent}: What is your location? \\ \hline
3 & \textbf{Customer}: Could you look for films shown in Napa? & \textbf{Customer}: Could you find films shown in San Ramon for me? \\ \hline
4 & \makecell{\textbf{Agent}: I discovered 1 movie - would you like Dumbo?} & \makecell{\textbf{Agent}: What is your take on Breakthrough or Captain Marvel?} \\ \hline
5 & & \makecell{\textbf{Customer}: Please look for other films. \\ I would like to watch at The Lot City Center.} \\ \hline
6 & & \textbf{Agent}: What is your take on Hellboy, Little, or Missing Link? \\ \hline
7 & \makecell{\textbf{Customer}: I'd love Captain Marvel. When can I watch it? \\ I'd like to watch a regular show.} & \makecell{\textbf{Customer}: Little is the one for me. When can I watch it? \\ I'd like to watch it today.} \\ \hline
8 & \textbf{Agent}: What time would you like to watch it? & \\ \hline
9 & \textbf{Customer}: I'd like to watch it on the day after tomorrow. & \\ \hline
10 & \makecell{\textbf{Agent}: I discovered 1 showtime for the film in Century \\ Napa Valley and XD Theater at 10:30 pm.} & \makecell{\textbf{Agent}: I discovered 1 showtime for the film at 2:30 pm \\ in The Lot City Center.} \\ \hline
11 & \textbf{Customer}: That sounds wonderful; that's all. & \textbf{Customer}: That sounds perfect for me; that's all for now. \\ \hline
12 & \textbf{Agent}: Have a pleasant afternoon. & \textbf{Agent}: Have a pleasant afternoon. \\

\end{tabular}
}
\caption{Alignment of two sample conversations. Empty lines indicate the operations of insertion and deletion. Non-empty lines enumerated with the same index indicate  
utterances, subject for substitution.}

\label{tbl:conv-examples}
\end{table*}

\section{Evaluation and Results}
\label{sec-evaluation-results}

We evaluated the effectiveness of our model through two distinct approaches: intrinsic evaluation, assessing the ability of the model to capture dialog flow (Section~\ref{sec:intrinsic-evaluation}), and external human evaluation via crowd-sourced annotations (Section~\ref{sec:human-evaluation}). We compared our model to two competitive baselines used for estimating text similarity.

\subsection{Datasets}
\label{sec-datasets}

Two dialog datasets were used for evaluation: Schema-Guided Dialog (SGD) dataset  \citep{rastogi_towards_2020} and MSDialog \citep{infoseek2018qu}. SGD is a large corpus of task-oriented dialogs that were created by crowd-workers and 
follow pre-defined dialog skeletons. MSDialog is a real-world dialog dataset of question answering interactions collected from a forum for Microsoft products, where a subset of dialogs (over $2$K) was labeled with metadata, including dialog acts.

\subsection{Baseline Models}
\label{sec:baseline-models}
We selected two competitive baselines for the evaluation of document similarity assessment: (1) Universal Sentence Encoder, a common choice for generating sentence-level embeddings, where a document embedding is computed by averaging its individual sentence representations; and (2) doc2vec \citep{le_distributed_2014}, an embedding algorithm that generates a distributional representation of documents, regardless of their length. The latter has been shown to outperform other document embedding approaches \citep{lau-baldwin-2016-empirical, zhang2019evaluating}. The distance between two dialogs, \texttt{avgSemDist} and \texttt{d2vDist}, respectively, is then computed by the cosine similarity between the final dialog representations. For the \texttt{d2vDist} measure, we trained a \href{https://radimrehurek.com/gensim}{doc2vec implementation} on over $20$K SGD and $35$K MSdialog dialogs, respectively. Both models were trained with default parameter values for 40 epochs. After model training, individual document (dialog) representations were inferred using the pre-trained doc2vec models.

\subsection{Intrinsic Evaluation}
\label{sec:intrinsic-evaluation}
We next assess the key capability of the conversation similarity measure: the ability to capture conversation structure and its temporal flow.
Note that albeit our intrinsic evaluation is done against datasets that include labeled dialog acts, our novel method, \texttt{convED}, does not rely on those for computing the similarity between two conversations. The dialog acts are being used merely for evaluation purpose. Thus, the method can be applied to any conversational data, whether between two humans or between a bot and a human. It is also not restricted to specific participant roles and can handle multiple participants.

\paragraph{Dialog Structural Edit Distance (\texttt{structED})}
Both SGD and a subset of MSDialog are annotated with rich metadata, including \textit{acts} and \textit{slot names} (SGD), and \textit{intent type}, the equivalent of acts (MSDialog). For example, the agent utterance ``When would you like to check in?'' in the SGD corpus is labeled with an act of type \texttt{REQUEST} and a slot value of type \texttt{check\_in\_date}. Consequently, a dialog \textit{structure} for the flow of actions and corresponding slot values can be extracted using this metadata. While faithfully representing a dialog flow, this structural pattern does not reflect (albeit not completely agnostic to) the precise semantics of utterances underlying the acts 
-- a setup that offers a natural test-bed for evaluation of our similarity measure, compared to other methods.

Specifically, given a dialog, we define its \textit{action flow} as the temporal sequence of its dialog acts or intents, concatenated with alphabetically sorted slots where they exist. As a concrete example, utterance \#2 in conversation 1 in Table~\ref{tbl:conv-examples} would be represented as \texttt{[REQUEST\_location]}, and utterance \#10 in conversation 2 would be transformed into \texttt{[OFFER\_location,OFFER\_time]}.

For a conversation $c_i$, we denote the sequence of its dialog acts and slots by $da_i$. Note that within a certain domain, the set of possible dialog acts and slots spans a fixed set. Therefore, the traditional edit distance metric can be applied to assess the distance between the dialog act flows of two conversations. Adhering to the conventional approach, we define the cost of insertion and deletion as $1$, and the cost of substitution as $2$. The dialog structural edit distance (\texttt{structED}) between conversations $c_i$ and $c_j$ is then computed as the edit distance between the two sequences $da_i$ and $da_j$, normalized by the longest sequence length.

\paragraph{Correlation to \texttt{convED}}
We hypothesize that the pairwise conversation distance represented by the \texttt{convED} measure (Section~\ref{sec:conved}) will exhibit higher proximity to \texttt{structED}, than the distance computed by either of the baseline models. We tested this hypothesis by calculating the four measures on all distinct conversation-pairs ($c_i$, $c_j$), $i{\neq}j$, in a conversation set $\mathcal{C}$. 
We then computed Pearson's correlation between each of \{\texttt{convED}, \texttt{avgSemDist}, \texttt{d2vDist}\} and \texttt{structED}. Since \texttt{structED} carries over only little semantics, the highest correlation will be indicative of the measure that most faithfully captures the inter-dialog structural similarity.

We performed our evaluation on a subset of SGD dialogs in the domain of Events due to their diverse nature, and on the set of MSDialog conversations. Utilizing the bootstraping setup, we randomly sampled $100$ subsets of $200$ conversations, and averaged over individual sample correlations. Table~\ref{tbl:conv-correlation} summarizes the results. Evidently, \texttt{convED} outperforms the other baselines, exhibiting a higher mean correlation to \texttt{structED}.

\begin{table}[hbt]
\centering
\begin{tabular}{l|r|r}
dataset & SGD (Events)  & MSDialog \\ \hline
\# of dialogs & 871 \hspace{0.12in} & 35,000 \hspace{0.12in} \\ \hline
\texttt{avgSemDist} & 0.265 \hspace{0.12in} & 0.031 \hspace{0.12in} \\
\texttt{d2vDist} & 0.097 \hspace{0.12in} & 0.112 \hspace{0.12in} \\
\texttt{convED} & \textbf{0.540}** & \textbf{0.301}** \\

\end{tabular}

\vspace{-0.03in}
\caption{Mean Pearson's correlation between \texttt{structED} and the pairwise dialog distance computed using each model. The best result in a column is boldfaced. Significant \textit{differences} between \texttt{convED} and the two baselines are marked by ‘**’ (t-test, p<.001).}
\label{tbl:conv-correlation}
\end{table}

\paragraph{Ablation study}
We next ask to study the affect of alignment by actor type (Section~\ref{sec:conved}) on the evaluation results (Table~\ref{tbl:conv-correlation}). Relaxing the constraint of the alignment, and, thereby using Equation~\ref{eq:sub-cosine} for computation of the substitution cost, resulted in the correlation of $0.538$ for SGD (Events) and $0.267$ for MSDialog. While a considerable drop is evident for MSDialog, the SGD results remain practically unaffected. We attribute the latter result to the schematic nature of SGD dialogs, lacking naturalistic variation: customer and agent utterances follow a predefined pattern, and differ to an extent that prevents the algorithm to align (semantically-distant) cross-actor utterances, even if the same-actor alignment is not strictly imposed.

\subsection{Human Evaluation}
\label{sec:human-evaluation}

We further evaluated the \texttt{convED} measure by comparing it to the human perception of dialog similarity. We hypothesized that the suggested approach is likely to exhibit a higher agreement with human judgement, than the more competitive baseline \texttt{avgSemDist} (on the SGD data).

Rating the precise degree of similarity between two dialogs is an extremely challenging task due to the subjective nature of the relative perception of conversation similarity. Rather than directly estimating a similarity value through scale-based annotation, we cast the annotation task as a two-way comparison scenario. We presented the crowd with a conversation triplet: one \textit{anchor} and two \textit{candidate} conversations.
We used the \href{https://appen.com/}{Appen} annotation platform targeting only the highest quality workers, where $5$ annotators provided judgements for a sample of $500$ triplets. Appendix B contains guidelines for the annotation task.

\paragraph{Conversation Triplet Selection}
Our inspection of the data reveals that for $68.4\%$ of randomly selected triplets with an anchor and two candidate conversations $a, c_1, c_2$, the two methods -- \texttt{convED} and \texttt{avgSemDist} -- agree on their judgement of the relative similarity for $c_1$ and $c_2$ to $a$. This observation limits the potential benefits of the crowd-sourcing task that was designed to test which approach better resembles human judgement, when the two methods exhibit mutual-disagreement. We, therefore, adhere to the retrospective annotation paradigm, where triplets are selected in a way that the two approaches yield contrasting judgement on the relative similarity of conversations $c_1$ and $c_2$ to the anchor $a$. Appendix B presents an annotation triplet example.

\paragraph{Annotation Results}
We limited the crowd-sourcing evaluation to a subset of annotation examples with at least $80\%$ (4 out of 5) inter-annotator agreement; this resulted in $229$ samples out of $500$. Treating these high-confidence judgements as the ground truth, we computed the ratio of triplets that agree with human intuition, for each of the two methods: \texttt{convED} and \texttt{avgSemDist}. A higher ratio would indicate that the method -- \texttt{convED} or \texttt{avgSemDist} -- more closely resembling human judgment on dialog similarity. 
The evaluation yielded $73.3\%$ and $26.7\%$ agreement with human judgements, for \texttt{convED} and \texttt{avgSemDist}, respectively. This corroborated our hypothesis suggesting that our measure better captures human perception of dialog similarity.

\subsection{Runtime Considerations}
In this work we used a common dynamic programming algorithm implementation that computes edit distance between two sequences in quadratic time; implementation, sufficiently efficient for relatively short sequences, subject to alignment. Recall that semantic similarity between two utterances is calculated by computing utterance embeddings, followed by measuring their cosine similarity, where the former is the most time-consuming part of the flow. Caching pre-computed utterance representations results in efficient computation overall, where computing \texttt{ConvED} between two conversations takes $5\text{-}6$ms running on CPU. Uncached alternative results in nearly $200$ms for a dialog-pair.

\section{Conclusions and Future Work}
\label{sec-conclusions}

We presented a novel approach for measuring dialog similarity, capturing both conversation semantics and its structural properties. Our evaluation shows that this measure outperforms other baselines with respect to both intrinsic evaluation and agreement with human judgements.
The framework is easily adaptable to different settings by manipulating the cost functions. For example, addressing multi-party chat by assigning lower distance to utterances coming from parties that share the same role, or reducing the cost for general chit chat utterances, thus focusing on semantic similarity of the conversations' essence.  Our future work includes enhancements of the proposed measure with additional dialog-related traits, as well as its application to downstream tasks and adapting the framework to multiple conversation alignment.

\section{Ethical Considerations}
\label{ethical-consid}
We have collected crowd annotations using the Appen platform. Due to the task difficulty, the mean hourly rate was $100$\% higher than the US federal minimum wage and contributors were offered additional bonuses to incentivize high quality work. In a contributor satisfaction survey conducted by the platform, our pay was rated 4.3/5 and clarity of instructions was rated 4.8/5.

Contributors only provided answers for multiple choice questions, or selected text spans from presented dialogs, and did not create any new textual content. No identifiable information about the contributors will be released with the data.

\section*{Acknowledgements}
We are thankful to the anonymous reviewers for their constructive feedback. We are also grateful to Boaz Carmeli and Chani Sacharen for their helpful discussions and useful comments on the earlier versions of this work.

\bibliography{anthology,custom}
\bibliographystyle{acl_natbib}


\vspace{1cm}
\section*{Appendix A}

\subsection*{Optimization of the Scaling Factor $\alpha$}

One of the common definitions of the edit distance uses insertion ($w_{ins}$), deletion ($w_{del}$) and substitution ($w_{sub}$) weights of 1, 1 and 2, respectively. These weights, however, can vary according to the precise use case. As an example, a spelling correction system can extend the traditional edit distance weights according to the relative distance of the keys on the keyboard.
\\

In this work we define substitution weight to be proportional to the semantic similarity of two utterances, by multiplying the cosine distance between the corresponding utterance representations by a scaling factor $\alpha$. Recall that for semantic representations, $\cos(e_1^i, e_2^j)$ (and, therefore, also $1\text{--}\cos(e_1^i, e_2^j)$) yield values between $0$ and $1$. Considering cosine distance (by fixing $\alpha$ to $1$) as substitution cost in the below Equation would result in the situation where substitution (i.e., alignment) of two utterances is always inherently `cheaper' than insertion or deletion of utterances (whose cost is $1$), even though the latter are preferable in cases where no semantic relation exists between the two.

\begin{equation*}
\fontsize{10}{12}{ 
    w_{sub}(u_1^i, u_2^j) = \begin{cases} 
        \alpha{\times}(1\text{--}\cos(e_1^i, e_2^j)) \;\; \text{if } a_1^i == a_2^j \\
        \infty \hspace{15pt} \text{otherwise}
    \end{cases}
    }
    \label{eq:sub-cost}
\end{equation*}

As a concrete example, the conversation alignment between two conversations in Table 1 in the main paper, would result in the alignment presented in Table~\ref{tbl:conv-examples} below. Note the sub-optimal alignment of utterances \#5, \#6 and \#7 of the two conversations: while semantically distant, their substitution cost is lower than inserting gaps. The ultimate outcome of this situation is that fixing the scaling factor $\alpha$ to $1$ will result in alignment based solely on substitutions for two dialogs of the same length.
We, therefore, seek to find an `optimal' value for $\alpha$ -- the value that will yield a plausible alignment between pairs of dialogs. In this work we learn $\alpha$ by performing greed search over possible values in the $1\text{--}5$ range, with $0.1$ steps. We consider the optimal value to be the one that maximizes the correlation of \texttt{convED} with \texttt{structED} on a held-out set of $100$ conversations, that were excluded from further experiments. The values of $2.2$ and $2.7$ were assigned to $\alpha$ for SGD and MSDialog datasets, respectively.
\\

\begin{table*}[hbt]
\centering
\resizebox{\textwidth}{!}{
\begin{tabular}{l|l|l}
\# & Conversation 1  & Conversation 2 \\ \hline
1 & \makecell{\textbf{Customer}: I'd like to look for a film to watch. \\ I like adventure films.} & \textbf{Customer}: I'd like to search for a fun film to watch. \\ \hline
2 & \textbf{Agent}: Where are you located? & \textbf{Agent}: What is your location? \\ \hline
3 & \textbf{Customer}: Could you look for films shown in Napa? & \textbf{Customer}: Could you find films shown in San Ramon for me? \\ \hline
4 & \makecell{\textbf{Agent}: I discovered 1 movie - would you like Dumbo?} & \makecell{\textbf{Agent}: What is your take on Breakthrough or Captain Marvel?} \\ \hline
5 & \makecell{\textbf{Customer}: I'd love Captain Marvel. When can I watch it?} & \makecell{\textbf{Customer}: Please look for other films. \\ I would like to watch at The Lot City Center.} \\ \hline
6 & \textbf{Agent}: What time would you like to watch it? & \textbf{Agent}: What is your take on Hellboy, Little, or Missing Link? \\ \hline
7 & \textbf{Customer}: I'd like to watch it on the day after tomorrow. & \makecell{\textbf{Customer}: Little is the one for me. When can I watch it? \\ I'd like to watch it today.} \\ \hline
8 & \makecell{\textbf{Agent}: I discovered 1 showtime for the film in Century \\ Napa Valley and XD Theater at 10:30 pm.} & \makecell{\textbf{Agent}: I discovered 1 showtime for the film at 2:30 pm \\ in The Lot City Center.} \\ \hline
9 & \textbf{Customer}: That sounds wonderful; that's all. & \textbf{Customer}: That sounds perfect for me; that's all for now. \\ \hline
10 & \textbf{Agent}: Have a pleasant afternoon. & \textbf{Agent}: Have a pleasant afternoon. \\

\end{tabular}
}
\caption{Alignment of two sample conversations, fixing the substitution scaling factor to $1$. Utterance-pairs in lines \#5, \#6 and \#7 represent undesired substitutions (carrying over different semantics).}

\label{tbl:conv-examples}
\end{table*}

\section*{Appendix B -- Annotation Guidelines}

Below are the annotation guidelines supplied for our annotators in the crowd-sourcing experiment.

\subsection*{Goal}
The goal of this task is to assess the similarity between conversations in the domain of customer service. The purpose of conversations is to assist customers in obtaining information about music and sports events and making reservations. A typical conversation consists of multiple turns (interactions) between a customer and a human agent. Conversations normally follow a pre-defined structure with slight variations. As such, most interactions include queries about types, dates and locations of events, followed by booking tickets, and confirmation on the agent side.

\begin{figure}[!h]
    \centering
    \includegraphics[width=0.85\linewidth]{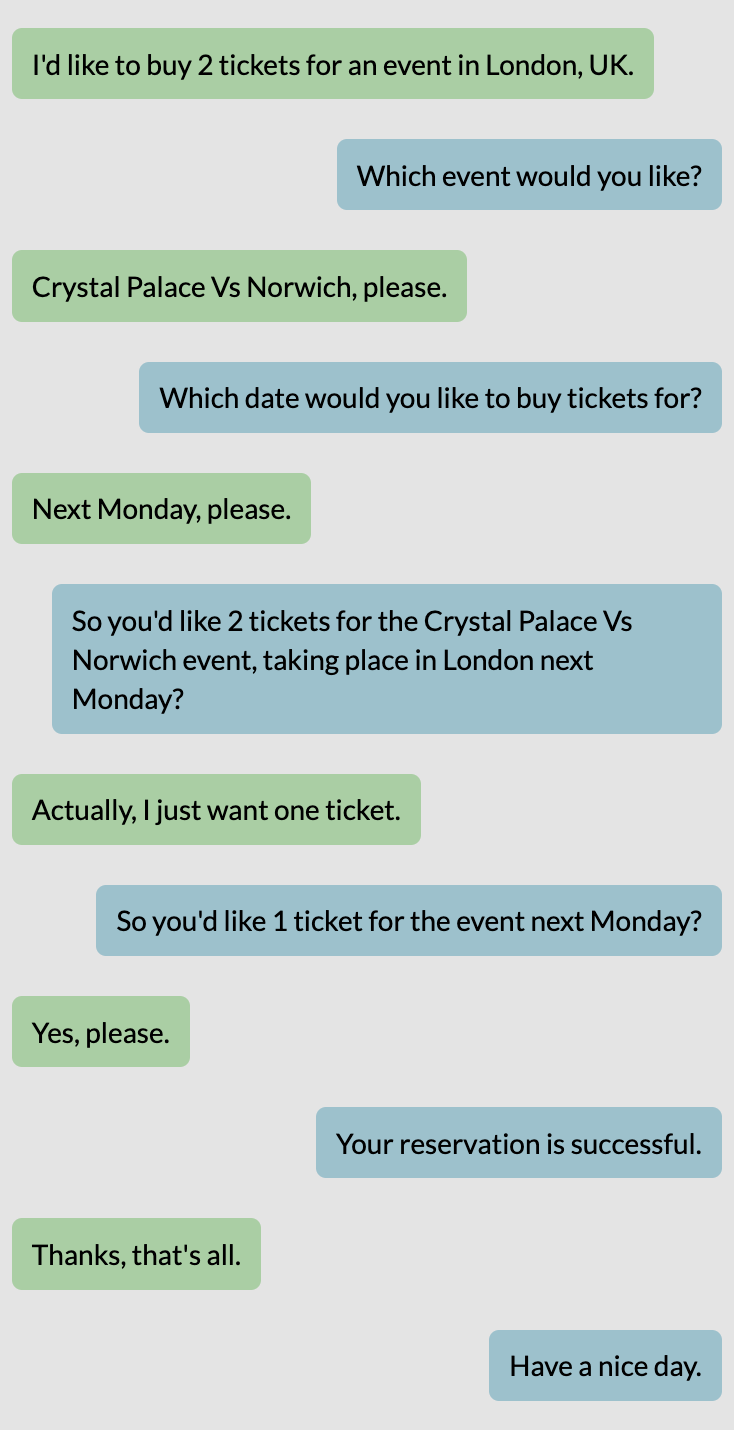}
    \caption{Example conversation from the dataset.}
    \label{fig:conv-example}
\end{figure}

Figure~\ref{fig:conv-example} presents an example conversation from the dataset. Conversations will often introduce some deviations from the depicted flow. As an example, customers may ask for a certain date or number of tickets, and then change their mind and details of their request, e.g., asking for a different number of tickets or another date. In some cases, customers only want to get information (date and time, location) about certain events, without actually making a reservation.

\subsection*{Rules and Tips}

In this task you will judge how similar conversations are, where similarity refers to multiple dimensions (the order of dimensions below does not necessarily imply relative importance):

\begin{itemize}[leftmargin=*]
\item \textit{Topical similarity}: how similar are the topics discussed, e.g., event type.
\item \textit{Conversation flow similarity}: how similar is the structure of conversations, e.g., interactions between the two actors, the final conversation outcome (e.g., reservation made or not).
\end{itemize}

Topics and structure are considered the major aspects that affect conversation similarity. While other details tend to vary between conversations, they carry over only minor (or no) effect on the judgement and, therefore should be ignored. As such, the precise event location or the name of the sports team selected should not affect the decision on conversation similarity. Consequently, two conversations that only differ in the following aspects should be considered identical: greetings and thanks (e.g., thank you for your help), precise sports team names or artist names (e.g., France Rocks Festival), precise locations and dates (e.g., tomorrow, 61 West 62nd street), precise number of tickets requested (e.g., 4 tickets).

Each annotation sample includes three conversations: the anchor conversation (in the middle, with grey background), and two candidate conversations -- one on the left-hand side and on the right-hand side. Your task is to decide which of the two candidate conversations is more similar to the anchor.
Note that both conversations can exhibit various extents of similarity to the anchor – spanning the whole range between very similar to completely distinct. Even if both candidates are seemingly very different from the anchor, your task is to still decide which of the two is more similar– the left or the right one.

\begin{figure*}[!h]
    \centering
    \includegraphics[width=1.0\linewidth]{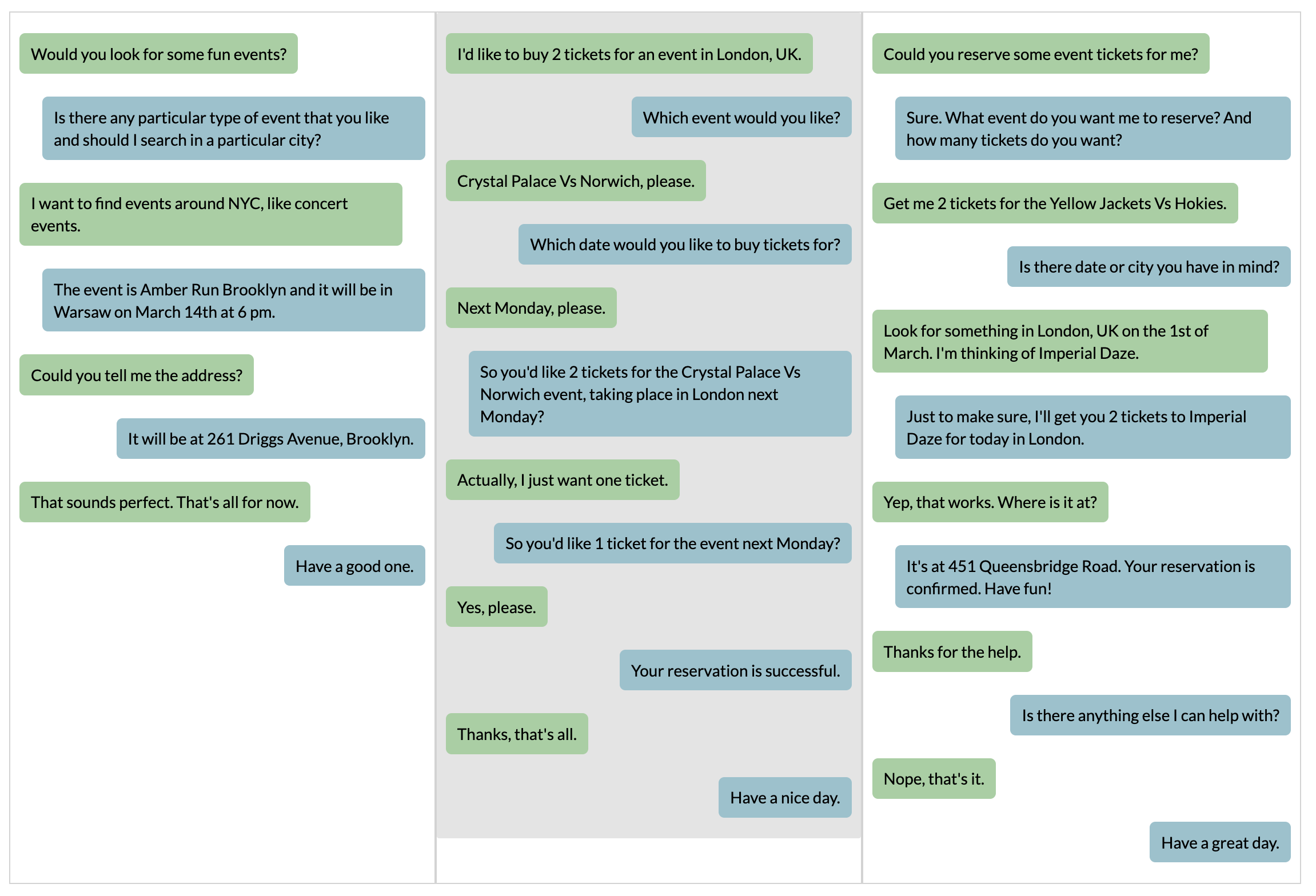}
    \caption{Example of an annotation item. The anchor conversation is in the middle (with grey background), while the two candidate conversations are on the left and on the right.}
    \label{fig:annotation-example}
\end{figure*}

\subsection*{Annotation Steps}

\begin{itemize}[leftmargin=*]
\item Read all three conversations carefully, beginning with the anchor (the middle conversation).

\item Answer the content question related to the anchor conversation. To answer the question, please select the relevant text section from the conversation and copy it to the answer text box. Some questions may entail answers that vary in their precise phrasing (e.g., "day after tomorrow", "the day after tomorrow"): select the phrasing you find most appropriate.

\item Think over the various aspects of similarity discussed above, and how they apply to your case. Note that the above guidelines leave some room for your intuition and (often subjective) judgement. However, you should be able to lay out the rationale behind each of your decisions.

\item Select the conversation more similar to the anchor between the two: click the radio button below -- either below the left hand-side conversation, or the right-hand side conversation.

\end{itemize}

We estimate the average time for each annotation example as a couple of minutes. Try to be decisive. In rare cases where you cannot decide between the two, click the “I can’t decide” radio button, located above the anchor conversation.

\subsection*{Examples}
As an example, consider the three conversations in Figure~\ref{fig:annotation-example}, where the middle one is the anchor.
\\

\noindent \textbf{Annotation answer}: the right-hand conversation is more similar to the anchor due to both higher topical similarity and structural similarity.

\end{document}